\documentclass{article}

\pdfoutput=1

\usepackage[nonatbib,preprint]{nips_2018}
\usepackage{hyperref}
\usepackage{url}

\usepackage[utf8]{inputenc} % allow utf-8 input
\usepackage[T1]{fontenc}    % use 8-bit T1 fonts
\usepackage{booktabs}       % professional-quality tables
\usepackage{multirow}       % multi-row in the table
\usepackage{amsfonts}       % blackboard math symbols
\usepackage{nicefrac}       % compact symbols for 1/2, etc.
\usepackage{microtype}      % microtypography

\usepackage{placeins}

\usepackage{graphicx}
\usepackage{color}
\usepackage{subcaption}
\usepackage{caption} 
\usepackage{algorithm}
\usepackage[noend]{algorithmic}

\usepackage{amsmath}
\usepackage{amssymb}
\usepackage{bm}
\usepackage{mathtools}
\usepackage{wrapfig}
\def\yes{$\surd$}

\newcommand{\kgenc}{\theta_{\text{enc}}}
\newcommand{\kgdec}{\theta_{\text{dec}}}
\newcommand{\prog}{\theta_{\text{prog}}}

\newcommand{\Hop}{\texttt{Pref}}
\newcommand{\HopFR}{\texttt{Suff}}
\newcommand{\Argmax}{\texttt{PrefMax}}
\newcommand{\ArgmaxFR}{\texttt{SuffMax}}

\newcommand{\ma}{a}

\newcommand{\mq}{\mathbf{q}}

\newcommand{\mc}{\mathbf{c}}
\newcommand{\wt}{text piece}
\newcommand{\mt}{\mathbf{t}}
\newcommand{\wT}{corpus}
\newcommand{\mT}{T}

\newcommand{\mProg}{\mathbf{p}}
\newcommand{\wprog}{statement}
\newcommand{\mprog}{p}

\newcommand{\wK}{knowledge storage}
\newcommand{\mK}{G}

\newcommand{\wk}{n-gram}
\newcommand{\mk}{\mathbf{g}}

\newcommand{\ms}{s}

\newcommand{\annot}[1]{$_\textrm{ #1}$}
\newcommand{\annt}[2]{{[#1]}$_\textrm{ #2}$}
\newcommand{\sym}[1]{{$\texttt{#1}$}}

\usepackage[utf8]{inputenc} % allow utf-8 input
\usepackage[T1]{fontenc}    % use 8-bit T1 fonts
\usepackage{hyperref}       % hyperlinks
\usepackage{url}            % simple URL typesetting
\usepackage{booktabs}       % professional-quality tables
\usepackage{amsfonts}       % blackboard math symbols
\usepackage{nicefrac}       % compact symbols for 1/2, etc.
\usepackage{microtype}      % microtypography

\title{Learning to Organize Knowledge 
and Answer Questions with N-Gram Machines}
\author{Fan Yang\thanks{\hspace{1mm} Part of the work was done while the author was interning at Google} 
\hspace{20pt} 
Jiazhong Nie 
\hspace{20pt} 
William W. Cohen 
\hspace{20pt} 
Ni Lao\thanks{\hspace{1mm} Part of the work was done while the author was working at Google}\\
Carnegie Mellon University, Pittsburgh, PA \\
Google LLC, Mountain View, CA \\
SayMosaic Inc., Palo Alto, CA \\
\texttt{\{fanyang1,wcohen\}@cs.cmu.edu},
\texttt{niejiazhong@google.com},
\texttt{ni.lao@mosaix.ai}
}

\begin{document}
\maketitle
\begin{abstract}
Though deep neural networks have great success in natural language processing,
%such as machine translation and reading comprehension, 
they are limited at more knowledge intensive AI tasks, such as open-domain Question Answering (QA).
Existing end-to-end deep QA models need to process the \emph{entire} text after observing the question, and therefore their complexity in responding a question is linear in the text size. 
This is prohibitive for practical tasks such as QA from Wikipedia, a novel, or the Web.
We propose to solve this scalability issue by using symbolic meaning representations, which can be indexed and retrieved efficiently with complexity that is independent of the text size.
%More specifically, we use sequence-to-sequence models to encode knowledge symbolically and generate programs to answer questions from the encoded knowledge.
%
%We apply our approach, called the N-Gram Machine (NGM), to the bAbI tasks~\cite{weston2015towards} and a special version of them (``life-long bAbI'') which has stories of up to 10 million sentences. 
%Our experiments show that NGM can successfully solve both of these tasks accurately and efficiently.
%Unlike fully differentiable memory models, NGM's time complexity and answering quality are not affected by the story length.
%
% The whole system of NGM is trained end-to-end with REINFORCE~\cite{williams1992simple}.
% To avoid high variance in gradient estimation, which is typical in discrete latent variable models, we use beam search instead of sampling.
% To tackle the exponentially large search space, we use a stabilized auto-encoding objective and a structure tweak procedure to iteratively reduce and refine the search space.
%Automatic ontology induction is a key yet unsolved problem in AI. 
%We conduct experiments on the \textsc{WikiMovies} dataset to test NGM's ability to induce an ontology from natural language text (Wikipedia) with only weak supervision (question-answer pairs), and correctly answer questions from the constructed ontology.
%
We apply our approach, called the N-Gram Machine (NGM), to three representative tasks.
First as proof-of-concept, we demonstrate that NGM successfully solves the bAbI tasks of synthetic text.
Second, we show that NGM scales to large corpus by experimenting on ``life-long bAbI'', a special version of bAbI that contains millions of sentences.
Lastly on the \textsc{WikiMovies} dataset, we use NGM to induce latent structure (i.e. schema) and answer questions from natural language Wikipedia text, with only QA pairs as weak supervision.
\end{abstract}

\section{Introduction}
Knowledge management and reasoning is an important task in Artificial Intelligence.
It involves organizing information in the environment into structured object (e.g. knowledge storage).
Moreover, the structured object is designed to enable complex querying by agents. 
In this paper, we focus on the case where information is represented in text.
An exemplar task is question answering from large corpus.
Traditionally, the study of knowledge management and reasoning is divided into independent subtasks, such as Information Extraction~\cite{dong2014vault,mitchell2015nell} and Semantic Parsing~\cite{dong2016language,jia2016data,liang2017nsm}.
Though great progress has been made on each \emph{individual} tasks, dividing the tasks upfront (i.e. designing the structure or schema) is costly, as it heavily relies on human experts, and sub-optimal, as it cannot adapt to the query statistics. 
To remove the bottleneck of dividing the task, \emph{end-to-end} models have been proposed for question answering, such as Memory Networks~\cite{miller2016key,weston2014memory}. 
However, these networks %Memory Networks 
lack scalability--
%More specifically, 
the complexity of reasoning with the learned memory is linear of the corpus size, which prohibits applying them to large web-scale corpus.

%TODO: (FY) following the logic in abstract, should first talk about scalability problem, then mention IE problem. 
%Although there is a great deal of recent research on extracting structured knowledge from text~\cite{dong2014vault,mitchell2015nell} and answering questions from structured knowledge stores~\cite{dong2016language,jia2016data,liang2017nsm}, much less progress has been made on either the problem of unifying these approaches in an end-to-end model or the problem of removing the bottleneck of relying on human experts to design the schema and annotate examples for information extraction. 
%In particular, traditional natural language processing and information extraction approaches are too labor-intensive and brittle for answering open domain questions from large corpus, and existing end-to-end deep QA models (e.g., \cite{miller2016key,weston2014memory}) lack scalability and the ability to integrate domain knowledge.
 
We present a new QA system that treats both the schema and the content of a structured storage as discrete hidden variables, and infers these structures  automatically from weak supervisions (such as QA pair examples).
The structured storage we consider is simply a set of ``n-grams'', which we show can represent a wide range of semantics, and can be indexed for efficient computations at scale. 
We present an end-to-end trainable system which combines a text auto-encoding component for encoding knowledge, and a memory enhanced sequence to sequence component for answering questions from the encoded knowledge. 
%We show that the method scales well on artificially generated stories of up to 10 million lines long (Figure~\ref{fig:compares}).
The system we present illustrates how end-to-end learning and scalability can be made possible through a symbolic knowledge storage.
% we probably should talk a bit more about MemNet

%\subsection{Question Answering As A Testbed for Text Understanding}
\subsection{Question Answering: Definition and Challenges}
\label{sec:qa}
We first define question answering as producing the answer $\ma$ given a corpus $\mT = \mt_1, \dots, \mt_{|\mT|}$, which is a sequence of \wt's, and a question $\mq$.
Both $\mt_i = \mt_{i,1}, \dots, \mt_{i,|\mt_i|}$  and  $\mq = \mq_1, \dots, \mq_{|\mq|}$ are sequences of words.
We focus on extractive question answering, where the answer $a$ is always a word in one of the sentences. 
In Section~\ref{sec:wikimovies} we illustrate how this assumption can be relaxed by named entity annotations.
Despite its simple form, question answering can be incredibly challenging.
We identify three main challenges in this process, which our new framework is designed to meet.
%There are three main challenges: % in this process:
%Now we present our new framework, which meets all three challenges described above.
%

\paragraph{Scalability}
A typical QA system, such as Watson~\cite{ferrucci2010watson} or any of the commercial search engines~\cite{brin1998search}, processes millions or even billions of documents for answering a question.
Yet the response time is restricted to a few seconds or even fraction of seconds.
Answering any possible question that is answerable from a large corpus with limited time means that the information need to be organized and indexed for fast access and reasoning.
% (FY:) comment out to save space
% Among all the challenges, scalability is the main blocker to a practical end-to-end solution. 

\paragraph{Representation} 
A fundamental building block for text understanding is paraphrasing. %, e.g., knowing that ``X married Y'' leads to ``X's wife is Y''.
Consider answering the question ``Who was Adam Smith's wife?' from the Web. 
There exists the following snippet from a reputable website 
\emph{``Smith was born in Kirkcaldy, ... (skipped 35 words) ... In 1720, he married Margaret Douglas''}.
An ideal system needs to identify that ``Smith'' in this text is equivalent to ``Adam Smith'' in the question; ``he'' is referencing ``Smith''; and text expressions of the form ``X married Y'' answer questions of the form ``Who was X's wife?''.

By observing users' interactions, a system may capture certain equivalence relationships among expressions in questions~\cite{baker2010understand}. 
However, given these observations, there is still a wide range of choices for how the meaning of expressions can be represented.
\emph{Open information extraction} approaches~\cite{angeli2015oie} represent expressions by themselves, and rely on corpus statistics to calculate their similarities. 
This approach leads to data sparsity, and brittleness on out-of-domain text.
\emph{Vector space} approaches~\cite{mikolov2013composition, weston2014memory, neelakantan2015neural, miller2016key} embeds text expressions into latent \emph{continuous} spaces. 
They allow flexible matching of semantics for arbitrary expressions, but are hard to scale to knowledge intensive tasks, which require inference with large amount of data.

\paragraph{Reasoning}
The essence of reasoning is to combine pieces of information together.
%The most basic form of reasoning is to combine pieces of information and form a new piece. 
For example, from \emph{co-reference}(``He'', ``Adam Smith'') and \emph{has\_spouse}(``He'', ``Margaret Douglas'') to  \emph{has\_spouse}(``Adam Smith'', ``Margaret Douglas''). 
As the number of relevant pieces grows, the search space grows exponentially -- making it a hard search problem~\cite{lao2011random}. 
%Good representations dramatically reduce the search space.
Since reasoning is closely \emph{coupled} with how the text meaning is stored, 
an optimal representation should be learned end-to-end (i.e. jointly) in the process of knowledge storing and reasoning.

\subsection{N-Gram Machines: A Scalable End-to-End Approach}
\label{sec:ngm}
\vspace{-0.1in}
%\begin{floatingfigure}[!hr]{3in}
\begin{wrapfigure}{r}{2.5in}
%\begin{figure}[!hbtp]
\centering
\includegraphics[width=2.5in]{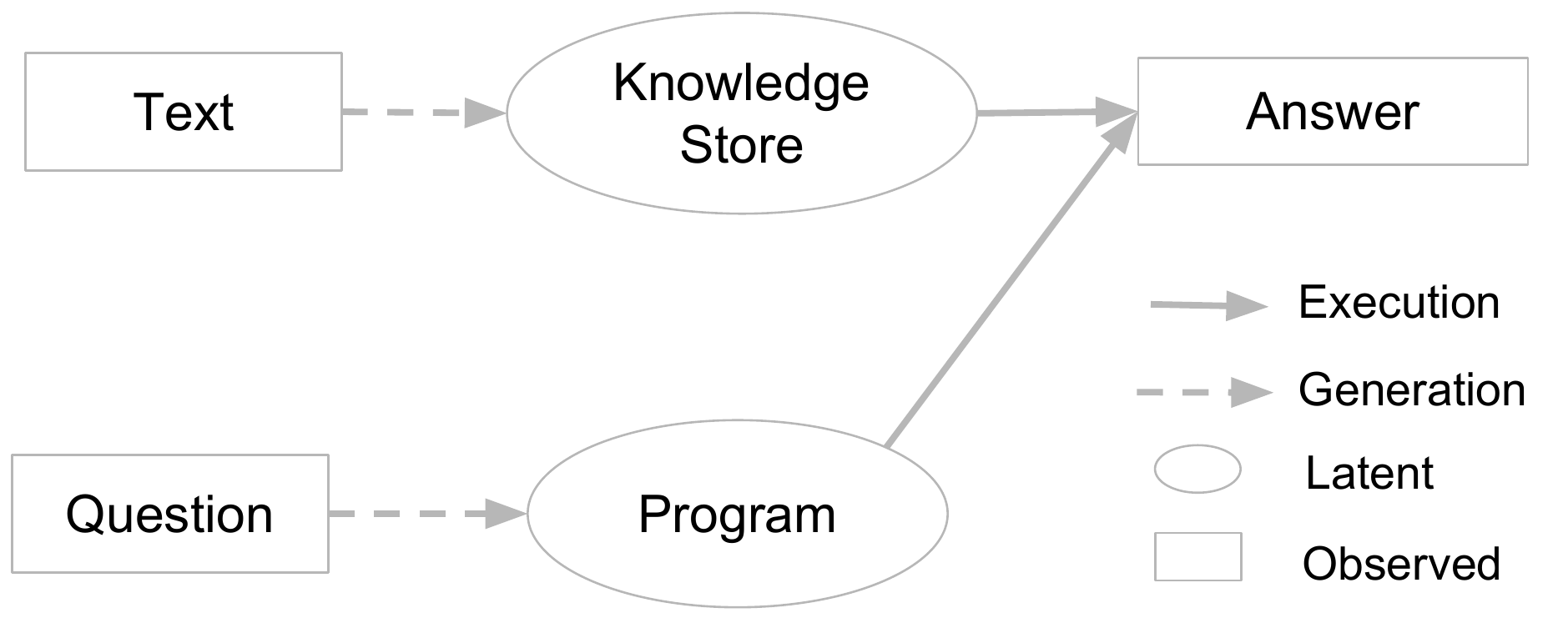}
\caption{An end-to-end QA system. 
%symbolic representation. 
Both the knowledge store and the program are non-differentiable and hidden.}
\label{fig:framework}
%\vspace{-0.1in}
%\end{floatingfigure}
\end{wrapfigure}

We propose to solve the scalability issue of neural network text understanding models by learning to represent the meaning of text as a \emph{symbolic} knowledge storage.
Because the storage can be indexed before being used for question answering, 
the inference step can be done very efficiently with complexity that is independent of the original text size.
More specifically the structured storage we consider is simply a set of ``n-grams'', which we show can represent complex semantics presented in bAbI tasks~\cite{weston2015towards} and can be indexed for efficient computations at scale. 
Each n-gram consists of a sequence of tokens, and each token can be a word, or any predefined special symbol.
Different from conventional n-grams, which are contiguous chunks of text, the ``n-grams'' considered here can be any combination of arbitrary words and symbols.
The whole system (Figure~\ref{fig:framework}) consists of learnable components which convert text into symbolic knowledge storage and questions into programs (details in Section~\ref{sec:model}).  
A deterministic executor executes the programs against the knowledge storage and produces answers. 
The whole system is trained end-to-end with no human annotation other than the expected answers to a set of question-text pairs.

\section{N-Gram Machines}
\label{sec:framework}
In this section we first describe the N-Gram Machine (NGM) model structure, which contains three sequence to sequence modules, and an executor that executes programs against knowledge storage.
Then we describe how this model can be trained end-to-end with reinforcement learning. 
We use the bAbI dataset~\cite{weston2015towards} as running examples.

\subsection{Model Structure}
\label{sec:model}
\paragraph{Knowledge storage}
Given a {\wT} $\mT = \mt_1, \dots, \mt_{|\mT|}$ NGM produces a {\wK} $\mK = \mk_1, \dots, \mk_{|\mT|}$, which is a list of {\wk}s.
An {\wk} $\mk_i=\mk_{i,1}, \dots, \mk_{i,N}$  is a sequence of symbols, % $\ms_1, \dots, \ms_N$, 
where each symbol $\mk_{i,j}$ is either a word from {\wt} $\mt_i$ or a symbol from the model vocabulary.
The {\wK} is probabilistic -- each {\wt} $\mt_i$ produces a distribution over {\wk}s, % $\mk_i$, 
and the probability of a {\wK} can be factorized as the product of {\wk} probabilities(Equation~\ref{eq:pkg}). 
Example {\wK}s are shown in Table~\ref{table:kg_example} and Table~\ref{tab:annotations}.
For certain tasks the model needs to reason over time. So we associate each {\wk} $\mk_i$ with a time stamp $\tau(\mk_i):=i$ with is simply its %the {\wt} 
id in {\wT} $\mT$.
%has two parts: a time stamp $i$ and 
%The time stamps in the tuple are useful for reasoning about time and are just sentence indices.

%\begin{table}[!htbp]
\begin{wraptable}{r}{3in}
\vspace{-0.2in}
\caption{Functions in NGMs. %N-Gram Machines. 
$\mK$ is the {\wK} (will be implicit in the actual program), and the input {\wk} is $\mathbf{v} = v_1, \dots, v_L$. We use $\mk_{i,:L}$ to denote the first $L$ symbols in $\mk_{i}$, and $\mk_{i,-L:}$ to denote the last $L$ symbols of $\mk_{i}$ in reverse order.
%a {\wk} $\mk_i$ is represented as $(\gamma_1, \dots, \gamma_N))$.
%``\texttt{FR}'' means \emph{from right}.
}
\label{table:functions}
\centering
\begin{tabular}{l}
\toprule
%Name  & Return \\
%\midrule
$\Hop(\mathbf{v}, \mK) =\{\mk_{i, L+1} \mid  \mk_{i,:L} = \mathbf{v}, \forall \mk_i \in \mK \}$ \\
$\HopFR (\mathbf{v}, \mK) =\{\mk_{i, -L-1} \mid  \mk_{i,-L:} = \mathbf{v}, \forall \mk_i \in \mK \}$ \\
$\Argmax (\mathbf{v}, \mK) =\{ \text{argmax}_{ g \in  \Hop(\mathbf{v}, \mK ) } \tau(g) \}$ \\
$\ArgmaxFR (\mathbf{v}, \mK) =\{ \text{argmax}_{ g \in \HopFR(\mathbf{v}, \mK )} \tau(g) \}$ \\
\bottomrule
\end{tabular}
\vspace{-0.1in}
%\end{table}
\end{wraptable}

\paragraph{Programs} 
The programs in NGM are similar to those introduced in Neural Symbolic Machines~\cite{liang2017nsm}, except that NGM functions operate on {\wk}s
%n-grams (i.e. knowledge tuples)\footnote{We will use ``n-gram'' and ``knowledge tuple'' interchangeably.} 
instead of Freebase triples.
NGM functions specify how symbols can be retrieved from a {\wK}
%Specifically, a function in NGM use a prefix (or  suffix) to retrieve symbols from tuples -- i.e. if a prefix ``matches'' a tuple, the immediate next symbol in the tuple is returned. 
%For the bAbI tasks, 
%We define four functions, which are illustrated
as  in  Table~\ref{table:functions}.
$\Hop$ and $\HopFR$ return symbols from all the matched {\wk}s, while $\Argmax$  and $\ArgmaxFR$  return  from the latest matches.
%(i.e. the {\wk} with the largest time stamp among all the matches).

More formally a program $\mProg$ is a list of {\wprog} $\mprog_1, ..., \mprog_{|\mProg|}$, where  $\mprog_i$ is either a special expression $\texttt{Return}$ indicating the end of the program, or is of the form  $f v_1 ... v_L$ where $f$ is a function in Table~\ref{table:functions} and $v_1 ... v_L$ are $L$ input arguments of $f$. 
When an expression is executed, it returns a set of symbols by matching its arguments in $\mK$, and stores the result in a new variable symbol (e.g., \sym{V1}) to reference the result (see Table~\ref{table:kg_example} for an example). 
Though executing a program on a {\wK} as described above is deterministic, probabilities are assigned to the execution results, which are the products of probabilities of the corresponding program and {\wK}. 
Since the knowledge storage can be indexed using data structures such as hash tables, the program execution time is independent of the size of the {\wK}.

\paragraph{Seq2Seq components}
NGM uses three sequence-to-sequence~\cite{sutskever2014sequence} neural network models to define probability distributions over {\wk}s and programs: 
%As  illustrated in Figure~\ref{fig:model}, these models are:
\begin{itemize}
\item A \textit{knowledge encoder} that converts {\wt}s to {\wk}s and defines a distribution $P(\mk|\mt, \mc; \kgenc)$.
It is also conditioned on context $\mc$ which helps to capture long range dependencies such as document title or co-references\footnote{Ideally it should condition on the partially constructed $\mK$ at time $i$, but that makes it hard to do LSTM batch training and is beyond the scope of this work.}.
The probability of a {\wK} $\mK = \mk_1 \dots \mk_{|\mT|}$ is defined as the product of its {\wk}s' probabilities:
\begin{align}
P(\mK|\mT;\kgenc) = \Pi_{\mk_i \in \mK} P(\mk_i|\mt_i, \mc_i;\kgenc)
\label{eq:pkg}
\end{align}
\item A \textit{knowledge decoder} that converts {\wk}s back to {\wt}s and defines a distribution $P(\mt|\mk, \mc;\kgdec)$.
It enables auto-encoding training, which is crucial for efficiently finding good knowledge representations (See Section~\ref{sec:opt}).
\item A \textit{programmer} that converts questions to programs and defines a distribution $P(\mProg|\mq,\mK;\prog)$. 
It is conditioned on the {\wK} $\mK$ for code assistance~\cite{liang2017nsm} -- before generating each token the programmer can query $\mK$ for valid next tokens given a n-gram prefix, and therefore avoid writing invalid programs.
\end{itemize}

%For all of these neural networks 
We use the CopyNet~\cite{gu2016incorporating} architecture, which has copy~\cite{vinyals2015pointer} and attention~\cite{bahdanau2014neural} mechanisms.
The programmer is also enhanced with a key-variable memory~\cite{liang2017nsm} for compositing semantics.

% \subsection{Decoding}
% Key-variable memory in NSM and multi-step reasoning.

% when the programmer decodes (given a question), it defines a distribution over all possible programs conditioned on the knowledge store of n-grams.

\subsection{Optimization}
%\subsection{Inference} 
\label{sec:search}
Given an example $(\mT, \mq, \ma)$ from the training set, NGM maximizes the expected reward
\begin{equation} O^{QA}(\kgenc, \prog)=\sum_{\mK} \sum_{\mProg} P(\mK|\mT;\kgenc) P(\mProg|\mq, \mK; \prog) R(\mK, \mProg, \ma), 
\label{eq:obj_slow}
\end{equation}
where the reward function $R(\cdot)$ returns 1 if executing $\mProg$ on $\mK$ produces $\ma$ and 0 otherwise. 
Since the training explores an exponentially large latent spaces, it is very challenging to optimize $O^{QA}$.
To reduce the variance of inference 
we approximate the expectations with beam searches instead of sampling. 
The summation over all programs is approximated by summing over programs found by a beam search according to $P(\mProg|\mq,\mK;\prog)$. 
For the summation over knowledge storages $\mK$, we first run beam search for each {\wt} based on  $P(\mk_i|\mt_i, \mc_i;\kgenc)$, and then sample a set of {\wK}s by independently sampling from the {\wk}s of each \wt. 
We further introduce two techniques to iteratively reduce and improve the search space:
\paragraph{Stabilized Auto-Encoding (AE)}
We add an auto-encoding objective to NGM, similar to the text summarization model proposed by Miao et al \cite{miao2016lang}.
The auto-encoding objective can be optimized by variational inference~\cite{kingma2014vae,mnih2014nvi}:
\begin{align}
O^{\text{VAE}}({\kgenc}, {\kgdec})
= \mathbb{E}_{p(z|x; {\kgenc})} [\log p(x|z; {\kgdec})   + \log p(z) - \log p(z|x; {\kgenc})],
\label{eq:vae_obj}
\end{align}
where $x$ is text, and $z$ is the hidden discrete structure.
However, it suffers from instability due to the strong coupling between encoder and decoder -- the training of the decoder ${\kgdec}$ relies solely on a distribution parameterized by the encoder ${\kgenc}$, which changes throughout the course of training.
To improve the %auto-encoding 
training stability, we propose to augment the decoder training with a more stable objective -- predict the data $x$ back from noisy partial observations of $x$, which are independent of ${\kgenc}$.
More specifically, for NGM we force the knowledge decoder to decode from a fixed set of hidden sequences $z \in \mathbf{Z}^N(x)$, which includes all {\wk} of length $N$ that consist only words from text $x$: 
\begin{align}
O^{\text{AE}}({\kgenc}, {\kgdec})
= \mathbb{E}_{p(z|x; {\kgenc})} [\log p(x|z; {\kgdec})  ] + \sum_{z \in \mathbf{Z}^N(x)} \log p(x|z; {\kgdec}),
\label{eq:stable_obj}
\end{align}
The knowledge decoder $\kgdec$ converts knowledge tuples back to sentences and the reconstruction log-likelihoods approximate how informative the tuples are, which can be used as reward for the knowledge encoder.
We also drop the KL divergence (last two terms in Equation~\ref{eq:vae_obj}) between language model $p(z)$ and the encoder, since the $z$'s are produced for NGM computations instead of human reading, and does not need to be in fluent natural language.
Sequential Denoising Autoencoder \cite{hill2016ul}.

  \begin{wrapfigure}{r}{2.7in}
  \vspace{-0.34in}
    \begin{minipage}{2.7in}
\begin{algorithm}[H] %[!htbp]
\caption{Structure tweak. 
%Here we assume that the function $f$ is one of \texttt{Hop} or \texttt{Argmax}. The procedure for \texttt{HopFR} and \texttt{ArgmaxFR} can be defined similarly.
}
\label{algo:tweak}
\begin{algorithmic}
\STATE{ 
{\bfseries Input:} 
{\wK} $\mK$; 
{\wprog} $f v_1 \dots v_L$ from an uninformed programmer. \\
\STATE{{\bfseries Initialize}  ${\mK}' = \emptyset$}
\IF{
$f(v_1 \dots v_L, \mK) \neq \emptyset$ or 
$f(v_1, \mK) = \emptyset$
%$v_1 \dots v_L$ can be matched in $\mK$, or $v_1$ can not be matched in $\mK$
}
	\RETURN
\ENDIF
\STATE{Let  $p=v_1 \dots v_m$ be the longest {\wk} prefix/suffix matched in $\mK$}
\STATE{Let  $\mK_p$ be the set of {\wk}s matching $p$.}
\FOR{$\mk = \ms_1 \dots \ms_N \in \mK_p$}{
\STATE{Add $v_1 \dots v_m v_{m+1} \ms_{m+2} \dots \ms_{N}$ to ${\mK}'$}
}\ENDFOR
{\bfseries Output} tweaked {\wk}s $\mK'$.}
\end{algorithmic}
\end{algorithm}
    \end{minipage}
  \end{wrapfigure}

\paragraph{Structure Tweaking (ST)}
% Even with AE training, the knowledge encoder is encoding tuples without the understanding of how they are going to be used, and may encode them inconsistently across sentences.
% At the later QA stage, such inconsistency can lead to no reward when the programmer tries to reason with multiple knowledge tuples. 
 NGM  contains two discrete hidden variables -- the {\wK} $\mK$, and the program $\mProg$. 
The training procedure only gets rewarded if these two representations agree on the symbols used to represent  certain concept (e.g., "X is the producer of a movie Y").
%
%To retrospectively correct the inconsistency in tuples,
To help exploring the huge search space more efficiently,
we apply \emph{structure tweak}, a procedure which is similar to code assist~\cite{liang2017nsm}, but works in an opposite direction --
while code assist uses the {\wK} to inform the programmer, structure tweak adjusts the knowledge encoder to cooperate with an uninformed programmer. 
Together they allow the decisions in one part of the model to be influence by the decisions from other parts -- similar to Gibbs sampling.  
%Markov Chain Monte Carlo. % (MCMC). 

More specifically, during training the programmer always performs an extra beam search with code assist turned off.
If the result programs lead to execution failure, 
the programs can be used to propose tweaked {\wk}s (Algorithm~\ref{algo:tweak}).
For example, when executing  
$\Hop$ \sym{john} \sym{journeyed} 
on the {\wK} in Table~\ref{table:kg_example}
%$\Hop$ \sym{mary} \sym{journeyed} on 
% a {\wK}   with tuples 
% (\texttt{john the milk}), 
% (\texttt{mary went bathroom}) and 
% (\texttt{mary picked milk}), 
matching the prefix \sym{john} \sym{journeyed} fails at symbol \sym{journeyed} and returns empty result.
At this point, %the programmer uses 
\sym{journeyed} can be used to replace inconsistent symbols in the partially matched {\wk}s (i.e. \sym{john to bedroom}), and produces  \sym{john journeyed bathroom}.
These tweaked tuples are then added into the  replay buffer for the knowledge encoder  ( Appendix~\ref{appx:opt}), %(Section~\ref{sec:opt}), 
which helps it to
%In this way, the search space of the knowledge storage is refined, and 
%the encoder can learn to %generate tuples using consistent symbols
adopt a vocabulary which is consistent with the programmer.
%in the future.  

%\subsection{Optimization}
\label{sec:opt}
Now the whole model has parameters $\theta=[\kgenc, \kgdec, \prog]$, and the training objective function is
\begin{align} 
O(\theta)
= \,& O^{AE}(\kgenc, \kgdec)  + O^{QA}(\kgenc, \prog) 
\label{eq:overall_obj}
\end{align}
Because the knowledge storage and the program are non-differentiable discrete structures,
we optimize our objective by a coordinate ascent approach -- optimizing the three components in alternation with REINFORCE~\cite{williams1992simple}.
See Appendix~\ref{appx:opt} for detailed training update rules.

\begin{wraptable}{rh}{2.8in}
%\begin{table}[!htbp]
%\scriptsize 
\footnotesize 
%\small
\vspace{-0.4in}
%\centering \textbf{
\begin{tabular}{ll}
\toprule
\textbf{Sentences} & \textbf{ %Time \&Encoded 
Knowledge Tuples} \\
%\cmidrule(lr){2-4} & Time stamp & Symbols & Probability \\
\midrule
Mary went to the kitchen &% 1, 
\texttt{mary to kitchen} \\ %, 0.9  \\
She picked up the milk & %2, 
\texttt{mary the milk}\\ %, 0.4 \\
John went to the bedroom & %3, 
\texttt{john to bedroom}\\ %, 0.7 \\
Mary journeyed to the garden & %4, 
\texttt{mary to garden}\\ %, 0.8 \\ 
\bottomrule
\end{tabular}
\caption{Example knowledge storage for bAbI tasks. 
%Each sentence may be converted to a distribution over multiple tuples, but only the one with the highest probability is shown here. 
To deal with coreference resolution we alway append the previous sentence as extra context to the left of the current sentence during encoding. 
%Consider executing the expression $\Hop$ \sym{V1} \sym{to} 
%against the {\wK} 
assuming that variable \sym{V1} stores \{\sym{mary}\} from previous executions.
%The execution 
Executing the expression $\Hop$ \sym{V1} \sym{to} 
returns a set of two symbols \{\sym{kitchen}, \sym{garden}\}.
Similarly, executing $\Argmax$ \sym{V1} \sym{to} would instead produces \{\sym{garden}\}. 
}
\label{table:kg_example}
%\vspace{-0.1in}
%\end{table}
\end{wraptable}

\section{bAbI Reasoning Tasks}
\label{sec:babi}
We apply the N-Gram Machine (NGM) to solve a set of text reasoning tasks in the Facebook bAbI dataset~\cite{weston2015towards}. 
%In Section~\ref{sec:babi}, 
We first demonstrate that the model can learn to build knowledge storage and generate programs that accurately answer the questions.
%In Section~\ref{sec:long_babi}, 
Then we show the scalability advantage of NGMs by applying it to longer stories %with large number of sentences 
up to 10 million sentences.

The Seq2Seq components are implemented as one-layer recurrent neural networks with Gated Recurrent Unit~\cite{chung2014empirical}.
The hidden dimension and the vocabulary embedding dimension are both 8.
We use beam size  2 for the knowledge encoder,  sample size  5 for the knowledge store, and  beam size 30 for the programmer.
%The neural networks are implemented in TensorFlow~\cite{abadi2016tensorflow} and optimized using Adam~\cite{kingma2014adam}.
%
We take a staged training procedure by first train with only the auto-encoding objective for 1k epochs, then add the question answering objective for 1k epochs, and finally add the structured tweak for 1k epochs. 
For all tasks, we set the {\wk} length to 3.

%\begin{table}[!htbp]
\begin{wraptable}{rh}{3.1in}
%\begin{floatingtable}[!hr]{
%\scriptsize 
%\footnotesize 
\small
%\centering
\vspace{-0.2in}
\begin{tabular}{lrrrrr}
\toprule
& T1 & T2 & T11 & T15 & T16 \\
\midrule
MemN2N &0.0 &83.0 &84.0 &0.0 &44.0 \\
QA &0.7 &2.7 &0.0 &0.0 &9.8 \\
QA + AE &70.9 &55.1 &\textbf{100.0} &24.6 &\textbf{100.0} \\
QA + AE + ST &\textbf{100.0} &\textbf{85.3} &\textbf{100.0} &\textbf{100.0} &\textbf{100.0} \\
\bottomrule
\end{tabular}
\caption{Test accuracy on bAbI tasks with auto-encoding (AE) and structure tweak (ST)
} \label{table:babi}
%\end{floatingtable}
%\end{table}
\end{wraptable}

\subsection{Extractive bAbI Tasks}

The bAbI dataset contains 20 tasks in total. 
We consider the subset of them that are extractive question answering tasks (as defined in Section\ref{sec:qa}).
Each task is learned separately.
%For all tasks, we set the knowledge tuple length to three.
%
In Table~\ref{table:babi}, we report results on the test sets.
NGM outperforms MemN2N~\cite{sukhbaatar2015end} on all tasks listed. 
The results show that auto-encoding is essential to bootstrap  learning--
without auto-encoding the expected rewards are near zero; but auto-encoding alone is not sufficient to achieve high rewards (See Section~\ref{sec:search}).
Since multiple discrete latent structures (i.e. knowledge tuples and programs) need to agree with each other over the choice of their representations for QA to succeed, the search becomes combinatorially hard.
Structure tweaking is an effective way to refine the search space -- improving the performance of more than half of the tasks.
Appendix \ref{appendix-babi} gives detailed analysis of auto-encoding and structure tweaks.

\subsection{Life-long bAbI}
\label{sec:long_babi}
To demonstrate the scalability advantage of NGM
%the N-Gram Machine, 
we conduct experiments on question answering from large synthetic \wT.
%data where the number of sentences may increase up to 10 million. 
More specifically we generated longer bAbI stories using the open-source script from Facebook\footnote{\url{https://github.com/facebook/bAbI-tasks}}.
We measure the answering time and answer quality of MemN2N~\cite{sukhbaatar2015end}\footnote{\url{https://github.com/domluna/memn2n}} and NGM at different scales.
The answering time is measured by the amount of time used to produce an answer when a question is given.
For MemN2N, this is the neural network inference time.
For NGM, because the knowledge storage can be built and indexed in advance, the response time  is dominated by LSTM decoding. 
%

%
%\begin{floatingfigure}[r]{3in}
\begin{wrapfigure}{r}{3in}
%\vspace{-0.1in}
%\begin{figure}[!hbtp]
\centering
\includegraphics[width=3in]{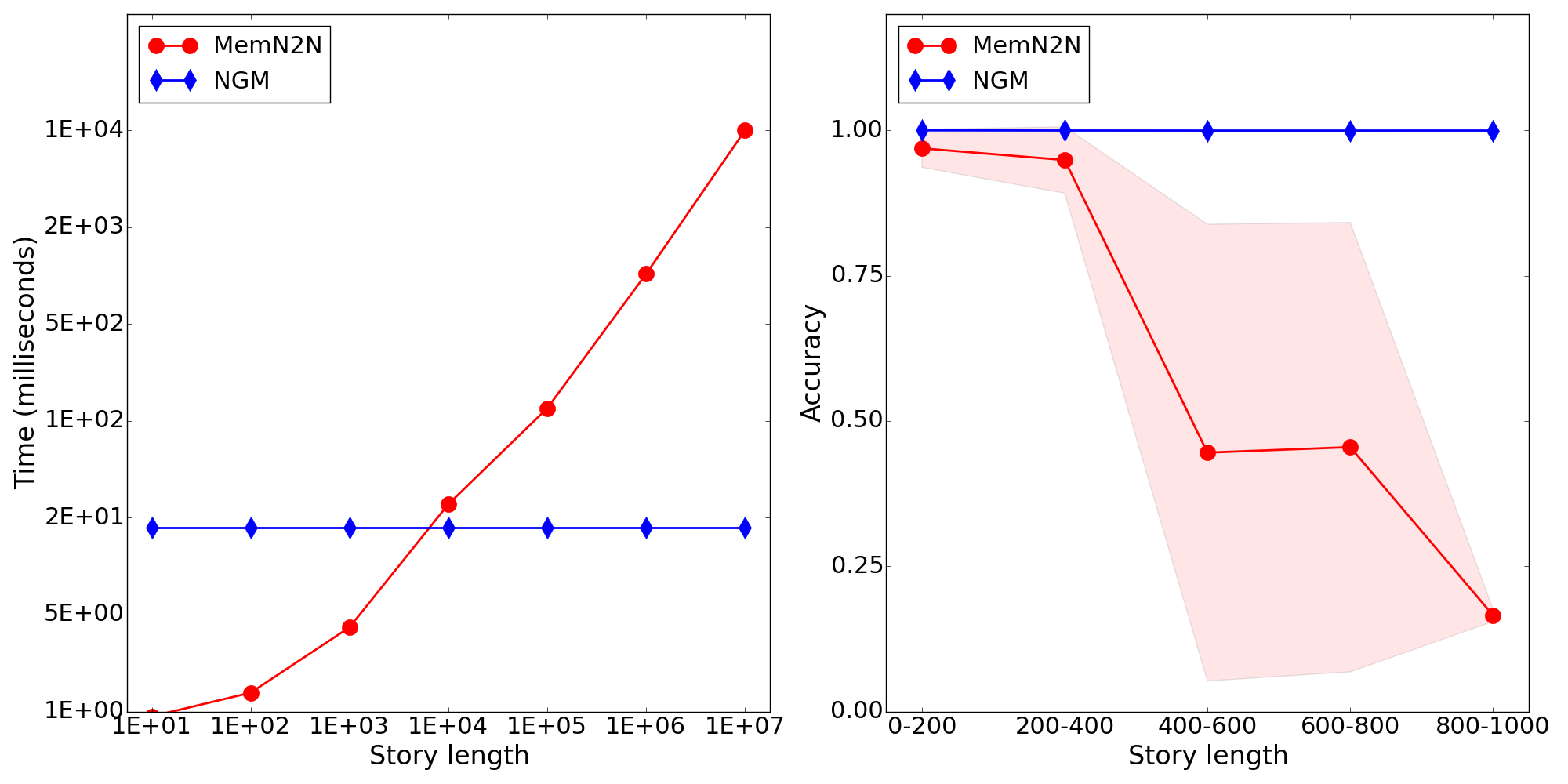}
\caption{Scalability comparison.
%of MemN2N and NGM. 
% Left: Answering time.
% Right: Answer quality.
Story length is the number of sentences in each QA pair.}
\label{fig:compares}
%\vspace{-0.5in}
%\end{figure}
\end{wrapfigure}
%\end{floatingfigure}
Figure~\ref{fig:compares} compares 
query response time of MemN2N and NGM.
% In terms of answering time, 
We can see that MemN2N scales poorly -- the inference time increases linearly as the story length increases.
 In comparison the answering time of NGM is not affected by story length.
\footnote{The crossover of the two lines is when the story length is around 1000, which is due to the difference in neural network architectures -- NGM uses recurrent networks while MemN2N uses feed-forward networks.}
%\nl{need to add index for MemN2N}
%
To compare the answer quality at scale, 
we apply MemN2N and NGM to solve three life-long bAbI tasks (Task 1, 2, and 11).
For each life-long task, MemN2N is run for 10 trials and the test accuracy of the trial with the best validation accuracy is used. 
For NGM, we use the same models trained on regular bAbI tasks.
We compute the average and standard deviation of test accuracy from these three tasks.
MemN2N performance is competitive with NGM when story length is no greater than 400, but decreases drastically when story length further increases.
On the other hand, NGM answering quality is the same for all story lengths.
These scalability advantages of NGM are due to its ``machine'' nature -- the symbolic knowledge storage can be computed and indexed in advance, and the program execution is robust on stories of various lengths.

% \section{Ontology Induction from Wikipedia}
\section{Schema Induction from Wikipedia}
\label{sec:wikimovies}

% (FY:) mentioned in Intro
%Automatic ontology induction is a key yet unsolved problem in AI.
We conduct experiments on the \textsc{WikiMovies} dataset to test NGM's ability to  induce an relatively simple schema from natural language text (Wikipedia) with only weak supervision (question-answer pairs), and correctly answer question from the constructed schema.

%\subsection{Problem Description}
The \textsc{WikiMovies} benchmark~\cite{miller2016key} %\footnote{\url{http://fb.ai/babi}} 
consists of question-answer pairs in the domain of movies.  
It is designed to compare the performance of various knowledge sources.
% (FY:) commented out to save space
%, including KB, IE, and document (Wikipedia) sources, when using a key-value memory network.
%It was built with the following goals in mind: (i) machine learning techniques should have ample training examples for learning; and (ii) one can analyze easily the performance of different representations of knowledge and break down the results by question type.
For this study we focus on the document QA setup, for which no predefined schema is given, and the learning algorithm is required to form an internal representation of the knowledge expressed in Wikipedia text in order to answer questions correctly.
It consists of 17k Wikipedia articles about movies. 
These questions are created ensuring that they are %all potentially 
answerable from the Wikipedia pages. 
In total there are more than 100,000 questions which fall into 13 general classes.
See Table~\ref{wikimovie} for an example document and related questions.
Following previous work \cite{miller2016key} we split the questions into disjoint training, development and test sets with 96k, 10k and 10k examples, respectively. 

\subsection{Text Representation}
The \textsc{WikiMovies} dataset 
%\textsc{WikiMovies}
comes with a list of entities (movie titles, dates, locations, persons, etc.), and we use Stanford CoreNLP~\cite{manning-EtAl:2014:P14-5} to annotate these entities in text.
Following previous practices in semantic parsing \cite{dong2016language,liang2017nsm} we leverage the annotations, and replaced named entity tokens with their tags for LSTM encoders, which significantly reduces the vocabulary size of LSTM models, and improves their generalization ability.
Different than those of the bAbI tasks, the sentences in Wikipedia are generally very long, and their semantics cannot fit into a single tuple.
Therefore, following the practices in~\cite{miller2016key}, instead of treating a full sentence as a \wt, we treat each annotated entity (we call it the anchor entity) plus a small window of 3 words in front of it as a \wt. 
%For each entity (besides the movie title) in the text, create a window where the first symbol is the movie title, the last symbol is the entity, and the symbols between are #window_size symbols that are not entities or punctuations. 
%We also find that adding the movie title as extra context is very helpful to capture long range dependencies, since many entities are directly related to the movie title.
We expect this small window to encode the relationship between the central entity (i.e. the movie title of the Wikipedia page) and the anchor entity. So we skip annotated entities in front of the anchor entity when creating the text windows.
We also append  the movie title as the context during encoding.
%
%Our representation is eventually very similar to the representation ``Window-level + Center Encoding + Title'' used in \cite{miller2016key}.
Table~\ref{tab:annotations} gives examples of the final document and query representation after annotation and window creation.
% For the question you listed, it requires 19 movies to be fully answerable, I just matched it to one of the movie.  
%We found that most of these entities have a direct relation to the central entity (a movie title), and these relations can answer most of the WikiMovies questions.

\begin{table}[t] \centering
\small
\begin{tabular}{p{0.95\linewidth}}
\hline
\textbf{Example document: Blade Runner} \\
\hline
Blade Runner is a 1982 American neo-noir dystopian science fiction film directed by Ridley Scott and starring Harrison Ford, Rutger Hauer, Sean Young, and Edward James Olmos. The screenplay, written by Hampton Fancher and David Peoples, is a modified film adaptation of the 1968 ... \\
\hline
\textbf{Example questions and answers} \\
\hline
Ridley Scott directed which films? Gladiator, Alien, Prometheus, Blade Runner, ... (19 films in total) \\
What year was the movie Blade Runner released? 1982\\
What films can be described by android? Blade Runner, A.I. Artificial Intelligence \\
%Who is the writer of the film Blade Runner?  \\
%... \\
\hline
\end{tabular}
\caption{Example document and question-answer pairs from the WikiMovies task. } \label{wikimovie}
\vspace{-0.1in}
\end{table}
\begin{table}[t] \centering 
%\scriptsize 
%\footnotesize 
\small
\begin{tabular}{l l}
\hline
\textbf{Annotated Text Windows(size=4): Blade Runner }  &  \textbf{Knowledge Tuples}  \\
\hline
%\annt{Blade Runner}{MOVIE}
is a [1982]\annot{DATE}&  [Blade Runner] when [1982] \\
%\annt{Blade Runner}{MOVIE}  
film directed by [Ridley Scott]\annot{PERSON}& [Blade Runner] director [Ridley Scott]\\
%\annt{Blade Runner}{MOVIE}
by and starring [Harrison Ford]\annot{PERSON}&  [Blade Runner] acted [Harrison Ford]\\
%\annt{Blade Runner}{MOVIE}
%by and starring [Rutger Hauer]\annot{PERSON} &[Blade Runner] acted [Rutger Hauer]\\
... & ... \\
%by and starring [sean young]\annot{PERSON} &[Blade Runner] acted [sean young]\\
%\annt{Blade Runner}{MOVIE}
and starring and [Edward James Olmos]\annot{PERSON} &[Blade Runner] acted [Edward James Olmos]\\
%\annt{Blade Runner}{MOVIE}
screenplay  written by [Hampton Fancher]\annot{PERSON} & [Blade Runner] writer [Hampton Fancher]\\
%\annt{Blade Runner}{MOVIE}
written by and  [David Peoples]\annot{PERSON}& [Blade Runner] writer [David Peoples]\\
\hline
\textbf{Annotated Questions} &   \textbf{Programs} \\
\hline
[Ridley Scott]\annot{PERSON} directed which films& 
$\HopFR$ [Ridley Scott] director  \\
What year was the movie [Blade Runner]\annot{MOVIE} released &
$\Hop$ [Blade Runner] when \\
What is movie written by [Hampton Fancher]\annot{PERSON} & 
$\HopFR$ [Hampton Fancher] writer \\
%What films can be described by [Android]\annot{MOVIE} & --\\
\hline
\end{tabular}
\caption{Annotated document and questions (left), and corresponding knowledge tuples and programs generated by NGM (right). %\footnote
%To help capturing the document wide context 
We always append the tagged movie title (\annt{Blade Runner}{MOVIE}) to the left of text windows as context during LSTM encoding.
%, and set its value to the movie title, i.e. "Blade Runner". } 
} \label{tab:annotations}
\vspace{-0.2in}
\end{table}

% \nl{Let's improve the result for Movie To Writer}
%\begin{table}[!htbp]
\begin{wraptable}{rh}{3.2in}
%\begin{floatingtable}[!hr]{
%\scriptsize 
%\footnotesize 
\small
%\centering
\vspace{-0.4in}
\begin{tabular}{lccccc}
\toprule
%QA + AE &70.9 &55.1 &\textbf{100.0} &24.6 &\textbf{100.0} \\
Question Type&	KB&	IE	&DOC&	NGM & U.B.\\
\midrule
Director To Movie&	0.90&	0.78&	0.91&	0.90&	0.91 \\
Writer To Movie&	0.97&	0.72&	0.91&	0.85&	0.89 \\
Actor To Movie&	0.93&	0.66&	0.83&	0.77&	0.86\\
Movie To Director&	0.93&	0.76&	0.79&	0.82&	0.91\\
Movie To Actors&	0.91&	0.64&	0.64&	0.63&	0.74\\
Movie To Writer&	0.95&	0.61&	0.64&	0.53&	0.86\\
Movie To Year&	0.95&	0.75&	0.89&	0.84&	0.92\\
\midrule
 Avg (extractive)& 0.93&	0.80&	0.70&	0.76 &0.87\\
\midrule
Movie To Genre&	0.97&	0.84&	0.86&	0.72&	0.75\\
Movie To Language&	0.96&	0.62&	0.84&	0.68&	0.74\\
Movie To Tags&	0.94&	0.47&	0.48&	0.43&	0.59\\
Tag To Movie&	0.85&	0.35&	0.49&	0.30&	0.59\\
Movie To Ratings&	0.94&	0.75&	0.92&	-&	- \\
Movie To  Votes&	0.92&	0.92&	0.92&	-&	- \\
\midrule
%Macro Average&	0.93&	0.65&	0.75&	0.68\\
%Micro Average&	0.94&	0.68&	0.76&	0.70\\	
 Avg (non-extractive)&	0.93&	0.75&	0.66&	0.36&	0.42 \\
\midrule
No schema design  &	&	&	\yes&	\yes&\\
No data curation  &	&\yes&	\yes&	\yes&\\
%No extraction training&	&	&\yes&	\yes&\\
Scalable inference&	\yes&\yes&	&	\yes&\\
\bottomrule
\end{tabular}
\caption%[Test accuracy]
{Scalability and test accuracy on WikiMovie tasks.\protect\footnotemark %\protect\footnotemark
%NGM is not applicable to IMDB rating or vote predictions, since the answers never appear in the Wikipedia page.
U.B. is the recall upper bound for NGM, which assumes that the answer appears in the relevant text, and has been identified by certain named entity annotator.
%\#Q is the number of queries in a query type.
} \label{table:wm-result}
\vspace{-0.4in}
%\end{floatingtable}
%\end{table}
\end{wraptable}
\footnotetext{We calculate macro average, which  is not weighted by the number of queries per type.} 
%\footnotetext{ }

%\subsection{Model and Training Details} 
%\subsection{Training and Testing Procedure}
\subsection{Experiment Setup}
Each training example  $(\mT, \mq, \ma)$ consists of $\mT$ the first paragraph of a movie Wikipedia page; a question $\mq$ from the \textsc{WikiMovies} for which its answer $\ma$ appears in the paragraph
\footnote{To speedup training we only consider the first answer of a question if there are multiple answers to this question. E.g., only consider ``Gladiator'' for the question ``Ridley Scott directed which films?''}.
We applied the same staged training procedure as we did for the bAbI tasks, but use LSTMs with larger capacities (with 200 dimensions). 100 dimension GloVe embeddings~\cite{Pennington2014GloveGV} are used as the input layer.
%, and the LSTMs have 200 dimension internal layers.
%
After training we apply  knowledge encoder to all  Wikipedia text with greedy decoding, and then aggregate all the {\wk}s into a single {\wK} $\mK^*$.
Then we apply  programmer with $\mK^*$ to every test question and greedy decode a program to execute and calculate F1 measure
%evaluate by 
using the %corresponding 
expected answers.

\subsection{Results}
We compare NGM with other approaches with different knowledge representations~\cite{miller2016key}.
%\nl{Summarize other approaches here.}
\textbf{KB} is the least scalable approach among all--needing human to provide both the schema and the contents of the structured knowledge.
\textbf{IE} is more scalable populating the contents of the structured knowledge, using information extractors pre-trained by human annotated examples.
\textbf{DOC} represents end-to-end deep models, which do not require any supervision other than question answer pairs, but are not scalable at answering time, because of the differentiable knowledge representations.
 
Table~\ref{table:wm-result} shows the performance of different approaches.
We separate the questions into two categories.
The first category consists of questions which are extractive --
assuming that the answer appears in the relevant Wikipedia text, and has been identified by Stanford CoreNLP\footnote{\url{https://stanfordnlp.github.io/CoreNLP/}} named entity annotator. 
NGM performance is comparable to IE and DOC approaches, but there is still a gap from the KB approach. This is because the NGM learned schema might not be perfect -- e.g., mixing writer and director as the same relation.
The second category consists of questions which are not extractive (e.g., for IMDB rating or vote predictions, the answers never appear in the Wikipedia page.), or we don't have a good entity annotator to identify potential answers. Here we implemented simple annotators for genre and language which have 59\% and 74\% coverage respectively. 
We don't have a tag annotator, but the tags can be partially covered by other annotators such as language or genre.
It remains as a challenging open question of how to expand NGM's capability to deal with non-extractive questions, and define {\wt}s without good coverage entity annotators.

\section{Related Work}
Training highly expressive discrete latent variable models on large datasets is a challenging problem due to the difficulties posed by inference \cite{hinton2006fla,mnih2014nvi}--specifically the huge variance in the gradient estimation.
Mnih et al\cite{mnih2014nvi} applies REINFORCE~\cite{williams1992simple} to optimize a variational lower-bound of the data log-likelihood, but relies on complex schemes to reduce variance in the gradient estimation.
We use a different set of techniques to learn N-Gram Machines, which are simpler and with less model assumptions. 
Instead of Monte Carlo integration, which is known for high variance and low data efficiency, we apply \emph{beam search}.
Beam search is very effective for deterministic environments with sparse reward~\cite{liang2017nsm, guu2017bridging}, but it leads to a search problem.
At inference time, since only a few top hypotheses are kept in the beam, search could get stuck and not receive any reward,
preventing learning.
We solve this hard search problem by having 1) a stabilized auto-encoding objective to bias the knowledge encoder to more interesting hypotheses; 
and 2) a structural tweak procedure which retrospectively corrects the inconsistency among multiple hypotheses so that reward can be achieved. %(See Section~\ref{sec:search}).

% The auto-encoding part of our model (Figure~\ref{fig:model}) is similar to the auto-encoding text summarization model proposed by \citet{miao2016lang}.
% \citet{miao2016lang} solved the large search space problem (when generating hidden sequences) by 
% 1) restricting hidden sequences to only consist of tokens in the source sequences (through a PointerNet~\cite{vinyals2015pointer}), and 
% 2) a language model pre-trained on a separate corpus.
% We adopt a different approach, which does not require a separate corpus, or a restricted hidden sequence space.
%
% 1) use a less restricted hidden space (through a CopyNet~\cite{gu2016incorporating}) by allowing both copied tokens and generated tokens;
% 2) stabilize the decoder by forcing it (through experience replay) to train from randomly generated hidden sequences; and
% 3) use the log-likelihood of the pre-trained decoder to guide the training of the encoder.

The question answering part of NGM our model (Figure~\ref{fig:model}) is similar to the Neural Symbolic Machine (NSM)~\cite{liang2017nsm}, which is a memory enhanced sequence-to-sequence model that translates  questions into programs in $\lambda$-calculus~\cite{liang11dcs}. The programs, when executed on a knowledge graph, can produce answers to the questions.
Our work extends NSM by removing the assumption of a given knowledge bases or schema, and instead learns to generate storage by end-to-end training to answer questions.

% In open-domain question answering, the retriever plays as important a role
% as the machine reader (Chen et al., 2017). In the
% past few years, there has been a lot of effort in designing sophisticated neural architectures for reading a small piece of text (e.g. paragraph) (Wang and Jiang, 2016; Xiong et al., 2016; Seo et al.,
%  2016; Lee et al., 2016, inter alia). However, most
%  work in open domain settings (Chen et al., 2017;
%  Clark and Gardner, 2017; Wang et al., 2018) only
% uses a simple retriever (such as TF-IDF based). As
%  a result, there is a notable decrease in the perfor-
%  mance of the QA system. A limitation for training a sophisticated retriever is the lack of available
%  training data which annotates the relevance of a
%  retrieved context with respect to the question. 
\section{Conclusion}
We present an end-to-end trainable system for efficiently answering questions from large corpus of text.
The system combines an text auto-encoding component for encoding the meaning of text into symbolic representations, and a memory enhanced sequence-to-sequence component that translates questions into programs.
We show that the method achieves good scaling properties and robust inference on syntactic and natural language text.
%artificially generated stories of up to 10 million sentences long.
%
The system we present here illustrates how a bottleneck in knowledge management and reasoning can be by alleviated by end-to-end learning of a symbolic knowledge storage.

% \input{discussion}

%\subsubsection*{Acknowledgments}

%\section*{References}

%\medskip
\FloatBarrier
\bibliographystyle{plain}
\bibliography{main}
%\bibliographystyle{nips_2018}
%\bibliographystyle{plainnat}
%\FloatBarrier

\newpage
\small
\appendix
\section{Supplementary Material}

\subsection{N-Gram Machines Details}
\label{appx:ngm}

\subsubsection{Model Structure Details}
\begin{figure}[!hbtp]
\centering
\includegraphics[width=5.5in]{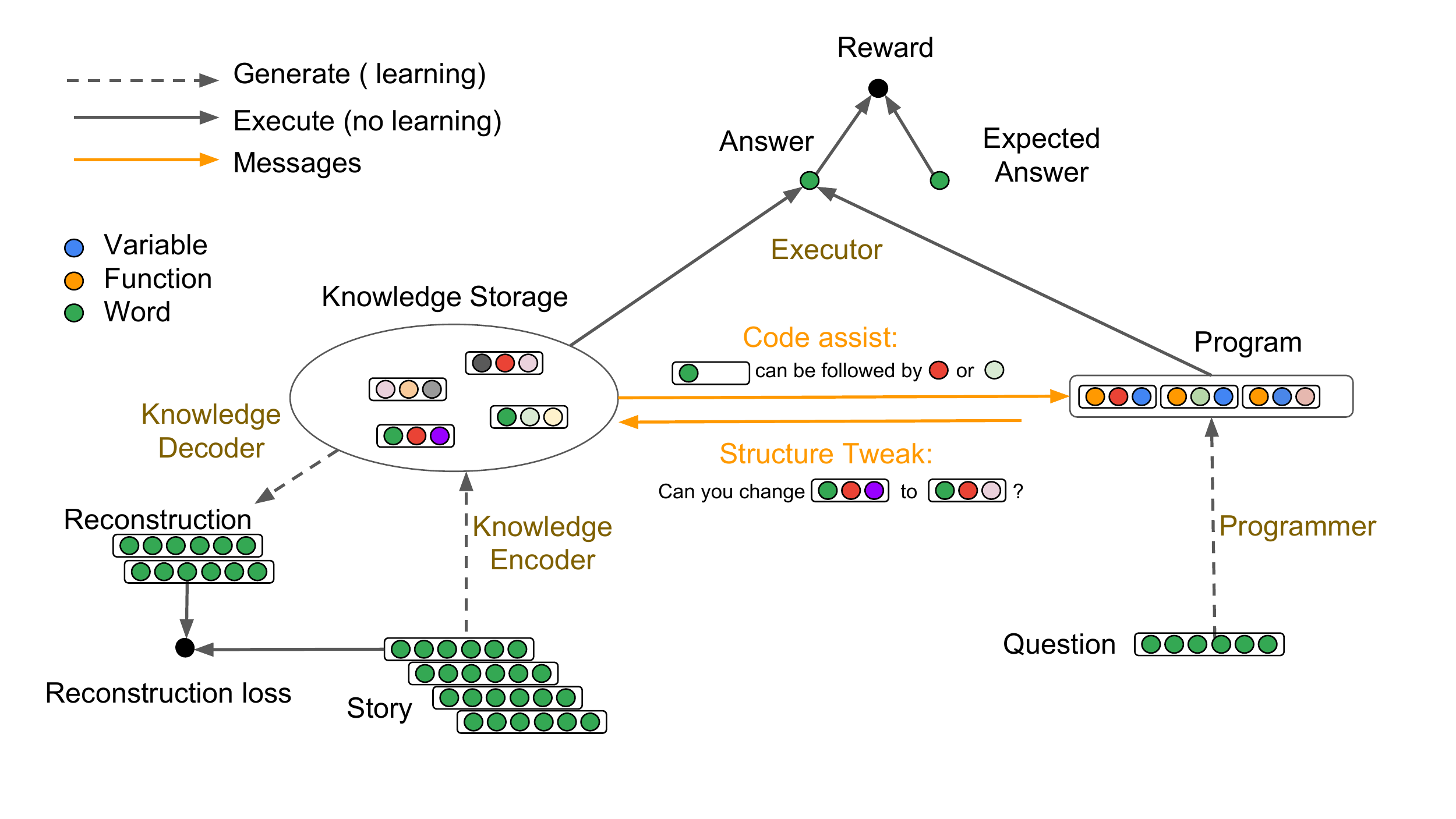}
\vspace{-0.2in}
\caption{N-Gram Machine. The model contains two discrete hidden structures, the knowledge storage and the program, which are generated from the story and the question respectively. 
The executor executes programs against the knowledge storage to produce answers.  
The three learnable components, knowledge encoder, knowledge decoder, and programmer, are trained to maximize the answer accuracy as well as minimize the reconstruction loss of the story.
Code assist and structure tweak help the knowledge encoder and programmer to communicate and cooperate with each other.}
\label{fig:model}
\end{figure}

Figure~\ref{fig:model} shows the overall model structure of an n-gram machine.

%P(\mK|\mT;\kgenc) = \Pi_{\mk_i \in \mK} P(\mk_i|\mt_i, \mc_i;\kgenc)
\subsubsection{Optimization Details}
\label{appx:opt}
The training objective function is
\begin{align} 
O(\theta)
= \,& O^{AE}(\kgenc, \kgdec)  + O^{QA}(\kgenc, \prog) \\
= \,& \sum_{i} \sum_{\mk_i}  [\beta(\mk_i) + P(\mk_i|\mt_i, \mc_i;\kgenc) ]  \log{P(\mt_i|\mk_i,\mc_i;\kgdec)}\\
& +\sum_{j} \sum_{\mK} \sum_{\mProg} P(\mK|\mT;\kgenc) P(\mProg|\mq^{(j)}, \mK; \prog) R(\mK, \mProg, \ma^{(j)})
% (FY:) commented out to avoid multiply defined label
%\label{eq:overall_obj}
\end{align}
where $i$ is the index to text unit and $j$ is the index to question answer pairs. $\beta(\mk_i)$ is 1 if $\mk_i$ only contains tokens from $\mt_i$ and 0 otherwise. 

For training stability and to overcome search failures, we augment this objective with experience replay~\cite{schaul2016prioritized}, and the gradients with respect to each set of parameters are: 
\begin{align} \label{eq:kgdec_gradient}
\nabla_{\kgdec} O'(\theta) = \sum_{i} \sum_{\mk_i} [\beta(\mk_i) + P(\mk_i|\mt_i, \mc_i;\kgenc) ] \nabla_{\kgdec} \log{P(\mt_i|\mk_i, \mc_i; \kgdec)},
\end{align}
\begin{align}
\nabla_{\kgenc} O'(\theta) 
= \,& \sum_{i} \sum_{\mk_i} 
[ P(\mk_i|\mt_i, \mc_i;\kgenc) \log{P(\mt_i|\mk_i, \mc_i;\kgdec)} \\
& + \mathcal{R}( \mathcal{G'}(\mk_i)) + \mathcal{R}( \mathcal{G}(\mk_i))
] \nabla_{\kgenc}{\log{P(\mk_i|\mt_i, \mc_i;\kgenc)}}  
\label{eq:kgenc_gradient},
\end{align}
where 
\begin{align}R( \mathcal{G}) = \sum_j \sum_{G\in \mathcal{G}} \sum_{\mProg} P(\mK|\mT;\kgenc) P(\mProg|\mq^{(j)}, \mK; \prog) R(\mK, \mProg, \ma^{(j)})\end{align}
is the total expected reward for a set of valid knowledge stores $\mathcal{G}$,
$\mathcal{G}(\mk_i)$ is the set of knowledge stores which contain the tuple $\mk_i$, and
$\mathcal{G'}(\mk_i)$ is the set of knowledge stores which contains the tuple $\mk_i$ through tweaking. 
\begin{align}
\nabla_{\prog} O'(\theta) 
= &\sum_j \sum_{G} \sum_{\mProg}
\left[\alpha I\left[\mProg \in \mathcal{B}^{(j)} \right] + P(\mProg|\mq^{(j)}, \mK; \prog)
\right] \\
& \cdot P(\mK|\mT;\kgenc) R(\mK, \mProg, \ma^{(j)})  \nabla_{\prog}{\log{P(\mProg|\mq^{(j)}, \mK; \prog)}},
\label{eq:prog_gradient}
\end{align}
where $\mathcal{B}^{(j)} $ is the experience replay buffer for $\mq^{(j)}$. $\alpha=0.1$ is a constant. 
During training, the program with the highest weighted reward (i.e. $P(\mK|\mT;\kgenc) R(\mK, \mProg, \ma^{(j)})$) is added to the replay buffer.

\subsection{Details of bAbI Tasks}
\label{appendix-babi}

\subsubsection{Details of auto-encoding and structured tweak}

\begin{figure}[!hbtp]
\caption{Visualization of the knowledge decoder's assessment of how informative the knowledge tuples are. Yellow means high and red means low.
}
\label{fig:vis_ae}
\centering
\includegraphics[width=0.8\linewidth]{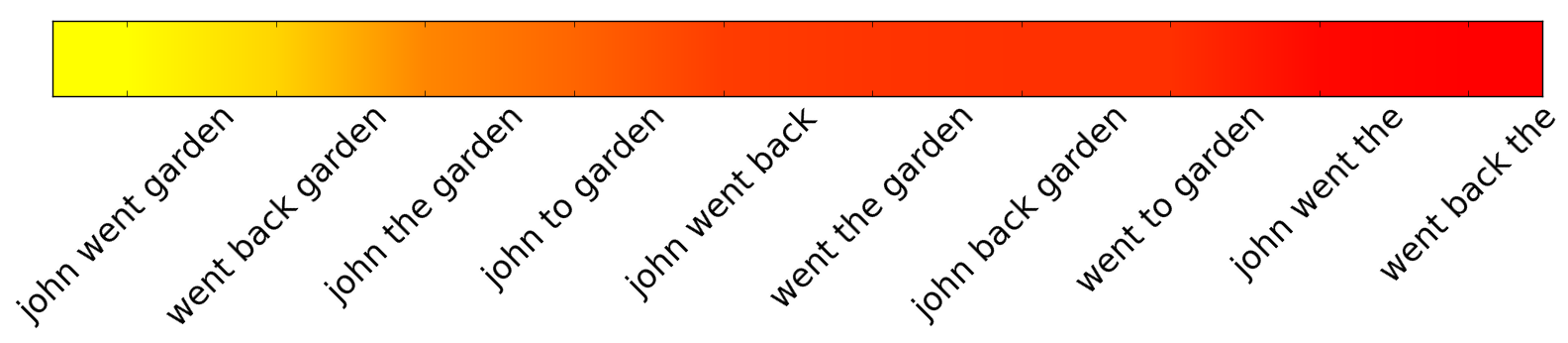}
\end{figure}

To illustrate the effect of auto-encoding, we show in Figure~\ref{fig:vis_ae} how informative the knowledge tuples are by computing the reconstruction log-likelihood using the knowledge decoder for the sentence "john went back to the garden". 
As expected, the tuple $(\texttt{john went garden})$ is the most informative.
Other informative tuples include $(\texttt{john the garden})$ and $(\texttt{john to garden})$.
Therefore, with auto-encoding training, useful hypotheses have large chance to be found by a small knowledge encoder beam size (2 in our case). 

Table~\ref{table:examples} lists sampled knowledge storages learned with different objectives and procedures. 
Knowledge storages learned with auto-encoding are much more informative compared to the ones without. 
After structure tweaking, the knowledge tuples converge to use more consistent symbols
 -- e.g., using \texttt{went} instead of \texttt{back} or \texttt{travelled}.
Our experiment results show the tweaking procedure can help NGM to deal with various linguistic phenomenons such as singular/plural (``cats'' vs ``cat'') and synonyms (``grabbed'' vs ``got'').
More examples are included in the supplementary material~\ref{sec:examples}.

\begin{table}[!htbp]
\caption{Sampled knowledge storage with question answering (QA) objective, auto-encoding (AE) objective, and structure tweak (ST) procedure. 
Using AE alone produces similar tuples to QA+AE.
The differences between the second and the third column are underlined.}
\label{table:examples}
\centering \small 
\setlength{\tabcolsep}{1pt}
\begin{tabular}{p{0.33\linewidth}p{0.33\linewidth}p{0.33\linewidth}}
\toprule
QA &QA + AE &QA + AE + ST \\
\midrule 
\texttt{went went went} &\texttt{daniel went office} &\texttt{daniel went office} \\
\texttt{mary mary mary} &\texttt{mary \underline{back} garden} &\texttt{mary \underline{went} garden} \\
\texttt{john john john} &\texttt{john \underline{back} kitchen} &\texttt{john \underline{went} kitchen} \\
\texttt{mary mary mary} &\texttt{mary \underline{grabbed} football} &\texttt{mary \underline{got} football} \\
\texttt{there there there} &\texttt{sandra got apple} &\texttt{sandra got apple} \\
\midrule
\texttt{cats cats cats} &\texttt{\underline{cats} afraid wolves} &\texttt{\underline{cat} afraid wolves} \\
\texttt{mice mice mice} &\texttt{\underline{mice} afraid wolves} &\texttt{\underline{mouse} afraid wolves} \\
\texttt{is is cat} &\texttt{gertrude is cat} &\texttt{gertrude is cat} \\
\bottomrule
\end{tabular}
\end{table}

\subsubsection{Model generated knowledge storages and programs for bAbI tasks} \label{sec:examples}
The following tables show one example solution for each type of task.
Only the tuple with the highest probability is shown for each sentence.

\begin{table}[!htbp]
\caption{Task 1 Single Supporting Fact}
\small
\centering 
\begin{tabular}{ll}
\toprule
\bf{Story} & \bf{Knowledge Storage}  \\
\midrule 
\texttt{Daniel travelled to the office.} 
& \texttt{Daniel went office} \\
\texttt{John moved to the bedroom.} 
& \texttt{John went bedroom} \\
\texttt{Sandra journeyed to the hallway.} 
& \texttt{Sandra went hallway} \\
\texttt{Mary travelled to the garden.} 
& \texttt{Mary went garden} \\
\texttt{John went back to the kitchen.} 
& \texttt{John went kitchen} \\
\texttt{Daniel went back to the hallway.} 
& \texttt{Daniel went hallway} \\
\midrule 
\bf{Question} & \bf{Program} \\
\midrule 
\texttt{Where is Daniel?} 
& \texttt{{\Argmax} Daniel went} \\
\bottomrule
\end{tabular}
\end{table}

\iffalse
\begin{table}[!htbp]
\caption{Task 2 Two Supporting Facts}
\small\centering
\begin{tabular}{ll}
\toprule
\bf{Story} & \bf{Knowledge Storage}  \\
\midrule 
\texttt{Sandra journeyed to the hallway.} & \texttt{Sandra journeyed hallway} \\
\texttt{John journeyed to the bathroom.} & \texttt{John journeyed bathroom} \\
\texttt{Sandra grabbed the football.} & \texttt{Sandra got football} \\
\texttt{Daniel travelled to the bedroom.} & \texttt{Daniel journeyed bedroom} \\
\texttt{John got the milk.} & \texttt{John got milk} \\
\texttt{John dropped the milk.} & \texttt{John got milk} \\
\midrule 
\bf{Question} & \bf{Program} \\
\midrule 
\texttt{Where is the milk?} 
& \texttt{{\ArgmaxFR} milk got} \\
& \texttt{{\Argmax} V1 journeyed} \\
\bottomrule
\end{tabular}
\end{table}
\fi

\begin{table}[!htbp]
\caption{Task 11 Basic Coreference}
\small\centering  
\begin{tabular}{ll}
\toprule
\bf{Story} & \bf{Knowledge Storage}  \\
\midrule 
\texttt{John went to the bathroom.} & \texttt{John went bathroom} \\
\texttt{After that he went back to the hallway.} & \texttt{John he hallway} \\
\texttt{Sandra journeyed to the bedroom} & \texttt{Sandra Sandra bedroom} \\
\texttt{After that she moved to the garden} & \texttt{Sandra she garden} \\
\midrule 
\bf{Question} & \bf{Program} \\
\midrule 
\texttt{Where is Sandra?} 
& \texttt{{\Argmax} Sandra she} \\
\bottomrule
\end{tabular}
\end{table}

\begin{table}[!htbp]
\caption{Task 15 Basic Deduction}
\small\centering 
\begin{tabular}{ll}
\toprule
\bf{Story} & \bf{Knowledge Storage}  \\
\midrule 
\texttt{Sheep are afraid of cats.} & \texttt{Sheep afraid cats} \\
\texttt{Cats are afraid of wolves.} & \texttt{Cat afraid wolves} \\
\texttt{Jessica is a sheep.} & \texttt{Jessica is sheep} \\
\texttt{Mice are afraid of sheep.} & \texttt{Mouse afraid sheep} \\
\texttt{Wolves are afraid of mice.} & \texttt{Wolf afraid mice} \\
\texttt{Emily is a sheep.} & \texttt{Emily is sheep} \\
\texttt{Winona is a wolf.} & \texttt{Winona is wolf} \\
\texttt{Gertrude is a mouse.} & \texttt{Gertrude is mouse} \\
\midrule 
\bf{Question} & \bf{Program} \\
\midrule 
\texttt{What is Emily afraid of?} 
& \texttt{{\Hop} Emily is} \\
& \texttt{{\Hop} V1 afraid}\\
\bottomrule
\end{tabular}
\end{table}

\begin{table}[!htbp]
\caption{Task 16 Basic Induction}
\small\centering  
\begin{tabular}{ll}
\toprule
\bf{Story} & \bf{Knowledge Storage}  \\
\midrule 
\texttt{Berhard is a rhino.} & \texttt{Bernhard a rhino} \\
\texttt{Lily is a swan.} & \texttt{Lily a swan} \\
\texttt{Julius is a swan.} & \texttt{Julius a swan} \\
\texttt{Lily is white.} & \texttt{Lily is white} \\
\texttt{Greg is a rhino.} & \texttt{Greg a rhino} \\
\texttt{Julius is white.} & \texttt{Julius is white} \\
\texttt{Brian is a lion.} & \texttt{Brian a lion} \\
\texttt{Bernhard is gray.} & \texttt{Bernhard is gray} \\
\texttt{Brian is yellow.} & \texttt{Brian is yellow} \\
\midrule 
\bf{Question} & \bf{Program} \\
\midrule 
\texttt{What color is Greg?} 
& \texttt{{\Hop}    Greg  a} \\
& \texttt{{\HopFR}  V1  a}\\
& \texttt{{\Hop}    V2  is}\\
\bottomrule
\end{tabular}
\end{table}

\end{document}